\documentclass[lettersize,journal]{IEEEtran}
\usepackage{amsmath,amsfonts}
\usepackage{algorithmic}
\usepackage{algorithm}
\usepackage{array}
\usepackage{textcomp}
\usepackage{stfloats}
\usepackage{url}
\usepackage{verbatim}
\usepackage{graphicx}
\usepackage{cite}
\usepackage{subfigure}
\usepackage{amsmath}
\usepackage{multirow}
\usepackage{amsthm}
\usepackage{amssymb }
\usepackage{amsmath,tabularx}
\makeatother
\usepackage{xcolor}
\usepackage{multirow}
\usepackage{caption}
\usepackage{float}
\usepackage{soul}
\usepackage{graphicx}
\usepackage{textcomp}
\usepackage{wrapfig}
\usepackage{amsmath}
\usepackage{bm}
\usepackage{cite}
\usepackage{flushend}
\usepackage{subfigure}
\usepackage{multirow}
\usepackage{graphicx}
\graphicspath{ {./images/} }
\begin{document}
\title{CLSA: Contrastive Learning-based Survival Analysis for Popularity Prediction in\\ MEC Networks}
\author{Zohreh Hajiakhondi-Meybodi,~\IEEEmembership{Student Member,~IEEE}, Arash Mohammadi,~\IEEEmembership{Senior Member,~IEEE}, 
Jamshid Abouei,~\IEEEmembership{Senior Member,~IEEE}, and~Konstantinos~N.~Plataniotis,~\IEEEmembership{Fellow,~IEEE}
\thanks{Z. Hajiakhondi-Meybodi is with Electrical and Computer Engineering (ECE), Concordia University, Montreal, Canada. (E-mail: z\_hajiak@encs.concordia.ca). A. Mohammadi (corresponding author) is with Concordia Institute of Information Systems Engineering (CIISE), Concordia University, Montreal, Canada. (P: +1 (514) 848-2424 ext. 2712 F: +1 (514) 848-3171, E-mail: arash.mohammadi@concordia.ca).  J. Abouei was with the Department of Electrical and Computer Engineering, University of Toronto, Toronto, Canada. He is now with the Department of Electrical Engineering, Yazd University, Yazd 89195-741, Iran (E-mail: abouei@yazd.ac.ir). K.~N.~Plataniotis is with Electrical and Computer Engineering (ECE), University of Toronto, Toronto, Canada. (E-mail: kostas@ece.utoronto.ca).}

\thanks{This Project was partially supported by Department of National Defence's Innovation for Defence Excellence \& Security (IDEaS)
program, Canada.}}
\maketitle

\begin{abstract}
Mobile Edge Caching (MEC) integrated with Deep Neural Networks (DNNs) is an innovative technology with significant potential for the future generation of wireless networks, resulting in a considerable reduction in users' latency. The MEC network's effectiveness, however, heavily relies on its capacity to predict and dynamically update the storage of caching nodes with the most popular contents. To be effective, a DNN-based popularity prediction model needs to have the ability to understand the historical request patterns of content, including their temporal and spatial correlations. Existing state-of-the-art time-series DNN models capture the latter by simultaneously inputting the sequential request patterns of multiple contents to the network, considerably increasing the size of the input sample. This motivates us to address this challenge by proposing a DNN-based popularity prediction framework based on the idea of contrasting input samples against each other, designed for the Unmanned Aerial Vehicle (UAV)-aided MEC networks. Referred to as the Contrastive Learning-based Survival Analysis (CLSA), the proposed architecture consists of  a self-supervised Contrastive Learning (CL) model, where the temporal information of sequential requests is learned using a Long Short Term Memory (LSTM) network as the encoder of the CL architecture. Followed by a Survival Analysis (SA) network, the output of the proposed CLSA architecture is probabilities for each content's future popularity, which are then sorted in descending order to identify the Top-$K$ popular contents. Based on the simulation results, the proposed CLSA architecture outperforms its counterparts across the classification accuracy and cache-hit ratio.
\end{abstract}
\maketitle
\section{Introduction} \label{sec:introduction}
Mobile Edge Caching (MEC) is a promising technology empowering the Beyond Fifth-Generation (5G) wireless networks to cope with the unprecedented expansion of the world's mobile data traffic~\cite{Sheraz2021}. MEC networks bring multimedia content closer to the Internet of Things (IoT) devices by offering storage capacities at the edge of the network, resulting in lower communication latency. The edge network's storage capacity, however, is constricted, making it impractical to keep all contents there. Existing reactive caching schemes~\cite{Hajiakhondi2019, Giovanidis2016} identify the most popular contents based on the observed request, where the content will be cached after being requested. Users' preferences, however, dynamically change over time. As a result, identifying the required contents before being requested and storing them on edge devices is a crucial part of the MEC networks.

As a prominent approach, Deep Neural Networks (DNN)-based proactive caching schemes~\cite{Doan2018, Ale2019, Zhang2019, Lin2020, Zhong2020, Wu2019, Wang2019, Mou2019} have been introduced. These models predict the Top-$K$ popular contents in the upcoming time using the historical request patterns of contents. To further enhance the performance of these schemes,  several time-series learning models have recently been introduced~\cite{Hajiakhondi2021_ICC, Hajiakhondi2022_IoT, Hajiakhondi2022_Icassp} to capture the temporal information of historical requests. Additionally, there is a meaningful correlation between the historical patterns of popular/unpopular content requests, making it necessary to learn the spatial correlation, along with capturing the temporal dependencies. The paper aims to further advance this emerging field.

\vspace{.05in}
\noindent
\textbf{Literature Review:} Generally speaking, popularity prediction models can be classified into three groups, i.e., $(i)$ Statistical models~\cite{Marlin2011, Odic2013}, e.g., content-based filtering, item-to-item correlation systems, and collaborative filtering; $(ii)$ Machine Learning (ML)-based schemes~\cite{Abidi2020}, such as Random Forest (RF)~\cite{Mendez2008}, Generalized Linear Model (GLM)~\cite{Ng2019}, and Decision Tree (DT)~\cite{Kabra2011}, and; $(iii)$ DNN-based architectures~\cite{Hajiakhondi2021_ICC, Hajiakhondi2022_IoT, Hajiakhondi2022_Icassp, Doan2018, Ale2019, Zhang2019, Lin2020, Zhong2020, Wu2019, Wang2019, Mou2019, Fan2021, Rathore2019}, such as Vision Transformers~\cite{Hajiakhondi2021_ICC, Hajiakhondi2022_Icassp}, Transformers~\cite{Hajiakhondi2022_IoT}, Autoencoders~\cite{Lin2020}, Long Short Term Memory (LSTM)~\cite{Zhang2019, Mou2019}, and Convolutional Neural Network (CNN)~\cite{Tsai2018}. Despite all the advantages of existing statistical and ML-based techniques, there are several limitations that render them ineffective for real-time caching solutions. On the one hand, statistical models would not be time-efficient when there are a huge number of users/multimedia contents. On the other hand, the sparsity and cold-start issues that occur when insufficient data is provided about a new mobile user/multimedia content affect the performance of the statistical and ML models. These issues, however, can be resolved by using existing DNN-based models. Last but not least, DNN-based frameworks predict the popularity of contents utilizing the raw historical request patterns of contents. Therefore, there is no requirement for feature engineering or pre-processing. These encourage the researchers to concentrate on DNN-based frameworks for popularity prediction in MEC networks.

Leveraging the CNN for feature extraction, a content-aware popularity prediction framework was introduced in~\cite{Doan2018}. Using statistical clustering methods, contents were categorized into different groups based on users' preferences. Similarly, Ndikumana \textit{et al.}~\cite{Ndikumana2021} used CNN and Multi-Layer Perceptron (MLP) to predict users' features and the probabilities that contents will be requested. Tang \textit{et al.}~\cite{Tang2020} implemented a Deep Reinforcement Learning (DRL) framework to model users' requesting behaviors. Sadeghi \textit{et al.}~\cite{Sadeghi2019} introduced an adaptive caching framework in hierarchical networks, where a two-way interactive influence between the cloud and edge devices is represented by a DRL model. These models, however, are inapplicable in highly dynamic practical networks since they make the assumption that the content popularity will not alter over time. 

To tackle the aforementioned issue, the main focus of recent works has been shifted to developing time-series prediction models~\cite{Ale2019, Jiang2020} to process the sequential and time-variant historical request patterns of contents. For instance, LSTM has been widely deployed in recent works, including but not limited to~\cite{Zhang2019, Mou2019, Zhang2021} for predicting the number of users' requests in the upcoming time. Although LSTM is capable of learning long-term dependencies, the correlation between the current sample and the earlier ones degrades over time. Moreover, they are unable to capture the spatial correlation between multiple contents. Another time-series learning model is Transformer architectures~\cite{Vaswani2017}, where there is no need to analyze sequential data in the same order. In our previous works~\cite{Hajiakhondi2021_ICC, Hajiakhondi2022_Icassp}, we deployed Vision Transformer (ViT)  to predict the Top-$K$ popular contents with high accuracy. To simultaneously predict the popularity of multiple contents and capture the spatial correlation between different contents, $2$D images were created to be used as the input of the ViT architecture, where each column of the image was related to the requests pattern of one content. In another work~\cite{Hajiakhondi2022_IoT}, we deployed a multi-channel Transformer architecture, where the historical request pattern of each content is given to a channel. Consequently, the spatial correlation of multiple contents is captured. Although providing higher accuracies, the complexity of the learning models in~\cite{Hajiakhondi2021_ICC, Hajiakhondi2022_Icassp, Hajiakhondi2022_IoT} exponentially increases by increasing the number of contents (i.e., the number of columns of the input image in the ViT model~\cite{Hajiakhondi2021_ICC, Hajiakhondi2022_Icassp} and the number of channels in the Transformer architecture~\cite{Hajiakhondi2022_IoT}). Another issue we encountered with these models~\cite{Hajiakhondi2021_ICC, Hajiakhondi2022_Icassp, Hajiakhondi2022_IoT} is that the input samples lack the contextual information of users such as age, and gender, making it ineffective to precisely understand distinct users' interests. To tackle these issues, we focus on the incorporation of Contrastive Learning (CL)~\cite{Khac2020}, which is a self-supervised learning paradigm that has been widely used (recently) in the computer vision domain. Intuitively speaking, by capitalizing on a unique characteristic of the CL paradigm, i.e., learning common attributes and differences in input samples, there is no need to simultaneously input the request patterns of all contents to capture the spatial correlations.    

Survival Analysis (SA)~\cite{Katzman2018} is another line of research that has been designed to learn the relationship between the contextual information of different users requesting the same content, and the distribution of the first time that the content will be requested again in the future. While SA models are widely used in medical studies, they are also applied in recommender systems~\cite{Jing 2017} to predict users' preferences in the future. For example, Desirena \textit{et al.}~\cite{Desirena2019} introduced a survival neural network to better interact with customers by modeling non-linear user preferences. To the best of our knowledge, there has been no study conducted on the joint use of CL and SA models for popularity prediction in MEC networks. This necessitates an urgent quest to develop and design a new and innovative SA-based popularity prediction architecture, which is the focus of this paper.

\vspace{.05in}
\noindent
\textbf{Contributions:}
Motivated by the above discussion, we introduce the Contrastive Learning-based Survival Analysis (CLSA) caching framework with the application to the MEC networks. The proposed CLSA architecture is a time-series popularity prediction framework learning the real-time content placement from the historical contextual information of users requesting a content. The proposed CLSA architecture consists of Reconstruction Network (RN), CL, and SA blocks, with the following characteristics:
\begin{itemize}
\item To provide a better understanding of historical patterns of content requests, longitudinal measurements are incorporated into a SA model.  The output of the SA model is the probability that one content gets popular within a time window, which is then sorted in descending order to identify the Top-$K$ popular contents. Moreover, a self-supervised CL network is used, focusing on the differences between the request patterns of distinct contents. 
\item Within the CL network, a shared encoder (developed based on the LSTM model) is used to learn a meaningful representation of historical patterns of distinct contents  to capture the temporal information as well. Consequently, it is unnecessary to input multiple historical patterns of contents simultaneously into the CLSA architecture to capture the spatial content correlations. This results in a significant reduction in the input size of the learning model. Since a CL network needs positive and negative samples, two augmentation methods are employed to create positive samples. Finally, a decoder is employed in the RN block to recreate the original input sample from the augmented one.
\end{itemize}
The effectiveness of the proposed CLSA framework is evaluated through comprehensive studies on the real-trace multimedia request patterns, focusing on classification accuracy and cache-hit ratio.  Simulation results corroborate that the CLSA framework outperforms its counterparts in all the above-mentioned aspects.

The remainder of the paper is organized as follows: In Section~\ref{Sec:2}, the system model and the dataset are described. Section~\ref{CLSA} presents the proposed CLSA architecture. Simulation results are presented in Section~\ref{sec:4}. Finally, Section~\ref{sec:5} concludes the paper.

\setlength{\textfloatsep}{0pt}
\begin{figure} [t!]
\centering \includegraphics [scale = 0.3] {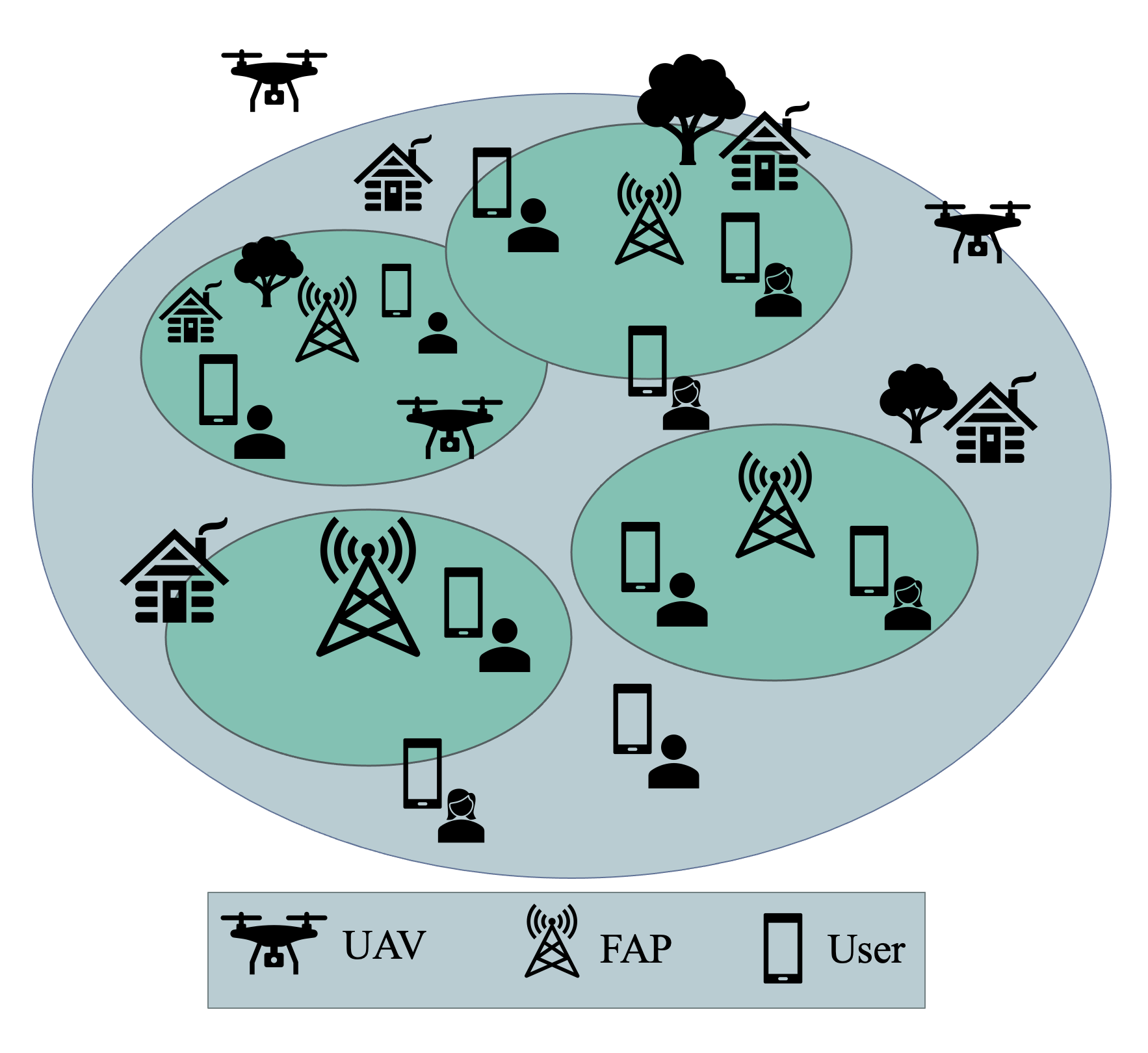}
\caption{\footnotesize A typical structure of the UAV-aided cellular network.} \label{sys}
\end{figure}

\section{System Model and Problem Description} \label{Sec:2}
We consider an Unmanned Aerial Vehicle (UAV)-aided MEC network, as shown in Fig.~\ref{sys}, which includes several UAVs as aerial caching nodes and Femto Access Points (FAPs) as terrestrial infrastructure. We define the set of UAVs as $u_{s}$, where $s \in \{1 , \ldots, N_u \}$, and the set of FAPs as $f_i$, with $i \in \{1 , \ldots, N_f \}$, serving requests of users, denoted by $u_l$, where $l \in \{1 , \ldots, U \}$. It is assumed that users request different multimedia contents, denoted by $c_m$, where $m \in \{1, \ldots, M \}$, belonging to a content library with the size of $M$. Moreover, UAVs and FAPs have a limited storage capacity. Following the common approach in the literature~\cite{Fadlullah2020}, it is assumed that all contents are the same size, and users only request one content at a time. In addition, FAPs and users are randomly distributed through the network with a Poisson Point Process (PPP) and Gaussian mixture distributions~\cite{Chen2017_2}, respectively, and UAVs' locations are obtained as the result of the $K$-Means clustering model~\cite{Hajiakhondi2021}. It is assumed that UAVs continue to hover over their locations as they fulfill a request for data delivery~\cite{Fadlullah2020}. The link quality and topology of UAVs are controlled by a Software Defined Network (SDN) controller, which also manages the aerial and terrestrial links~\cite{Fadlullah2020}.

\subsection{Dataset}
\begin{figure*} [t!]
\setlength{\textfloatsep}{0pt}
\centering \includegraphics [scale = 0.31] {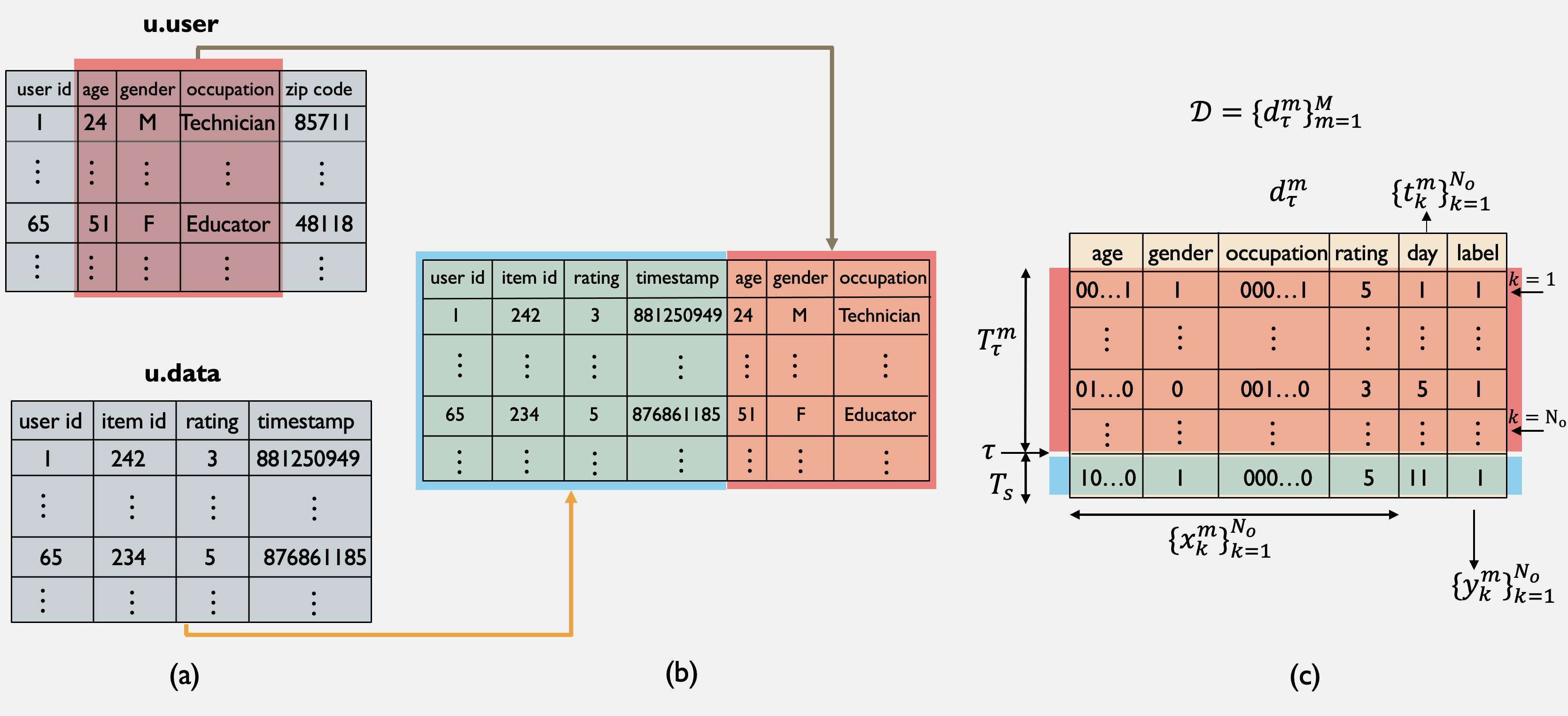}
\caption{\footnotesize (a) A typical sample of the Movielens dataset, (b) the concatenated dataset, and (c) the adopted version of the Movielens dataset used for the CLSA architecture.} \label{data}
\end{figure*}

MovieLens dataset~\cite{Harper2015}, as one of the most well-known movie recommendation services that provide users' contextual information, is used in this study to evaluate the proposed CLSA architecture. MovieLens was generated on October $17$, $2016$, including the movie rates that $943$ users gave to $1682$ movies between September $19$, $1997$, and April $22$, $1998$. As shown in Fig.~\ref{data}(a), \textit{u.user} document provided in the MovieLens dataset, contains  users' contextual information, including gender, age, occupation, and ZIP code. The ZIP codes are converted to latitude and longitude coordinates to extract users' locations during their requests. Another document provided in this dataset is \textit{u.data}, including user ID, item ID (content ID), the rate the user gave to the corresponding content, and the timestamp that the user watched and rated the content.

As shown in Fig.~\ref{data}, the following steps are performed to adopt the Movielens dataset with the CLSA architecture: $(i)$ The \textit{u.data} and \textit{u.user} documents are concatenated on ``user ID" (see Fig.~\ref{data}(b)) and the ZIP code is dropped from the concatenated dataset; $(ii)$ The concatenated dataset in Fig.~\ref{data}(b) is sorted by ``item ID", and ``timestamp"; $(iii)$ The ``user ID" and ``item ID" are dropped from the concatenated dataset, since these columns are not informative; $(iv)$ The categorical features including gender, age, and occupation are encoded using a one-hot encoder; $(v)$ The timestamp is discretized with a resolution of one day since the storage of edge devices should be updated during the off-peak times (i.e., midnight)~\cite{Vallero2020}. Then, the discretized timestamp, named ``day", is replaced with the timestamp column; and, $(vi)$ Finally, a column named ``label" is added to the concatenated dataset, indicating the content popularity (popular or unpopular), which will be described shortly.

We define an observational window for each content $c_m$, for ($1 \leq m \leq M$) having the length of $T_{\tau}^m$ at time $\tau$, where the request pattern of contents within this range are studied to predict their popularity in the future, i.e., the study window, denoted by $T_s$. Following the Reference~\cite{Hong2022}, we consider the same number of requests $N_o$ in the observational window for all the contents, where zero padding is used for contents with less number of requests. Given the request patterns of contents over the observational window at time $\tau$, the popularity of contents is predicted within the time window $[\tau, \tau + T_s ]$. Without loss of generality, since the storage capacity of edge devices should be updated each day, we assume that the length of the study window is $T_s = 1$. Given $M$ number of multimedia contents and $U$ number of users, the dataset $\mathcal{D} = \{ d_{\tau}^m \}^{M}_{m=1}$ is created, where $d_{\tau}^m = \{ (x_k^m, t_k^m,y_k^m ) \}^{N_o}_{k=1}$ is the time-series observational data for content $c_m$. Term $\{x_k^m\}^{N_o}_{k=1}$ includes the contextual information of users requesting content $c_m$ and the rating that these users gave to content $c_m$ (see Fig.~\ref{data}(c)).  Moreover, rating time is represented by $\{t_k^m\}^{N_o}_{k=1}$. Finally, term $y_k^m \in \{0,1\}$ is associated with the content popularity, which is $1$ if content $c_m$ gets popular during the study window $T_s$, otherwise, $y_k^m = 0$. Consequently, given $d_{\tau}^m$, the output of the proposed CLSA model predicts whether or not content $c_m$ will be popular during the study window $T_s$.

\begin{figure*} [t!]
\centering \includegraphics [scale = 0.35] {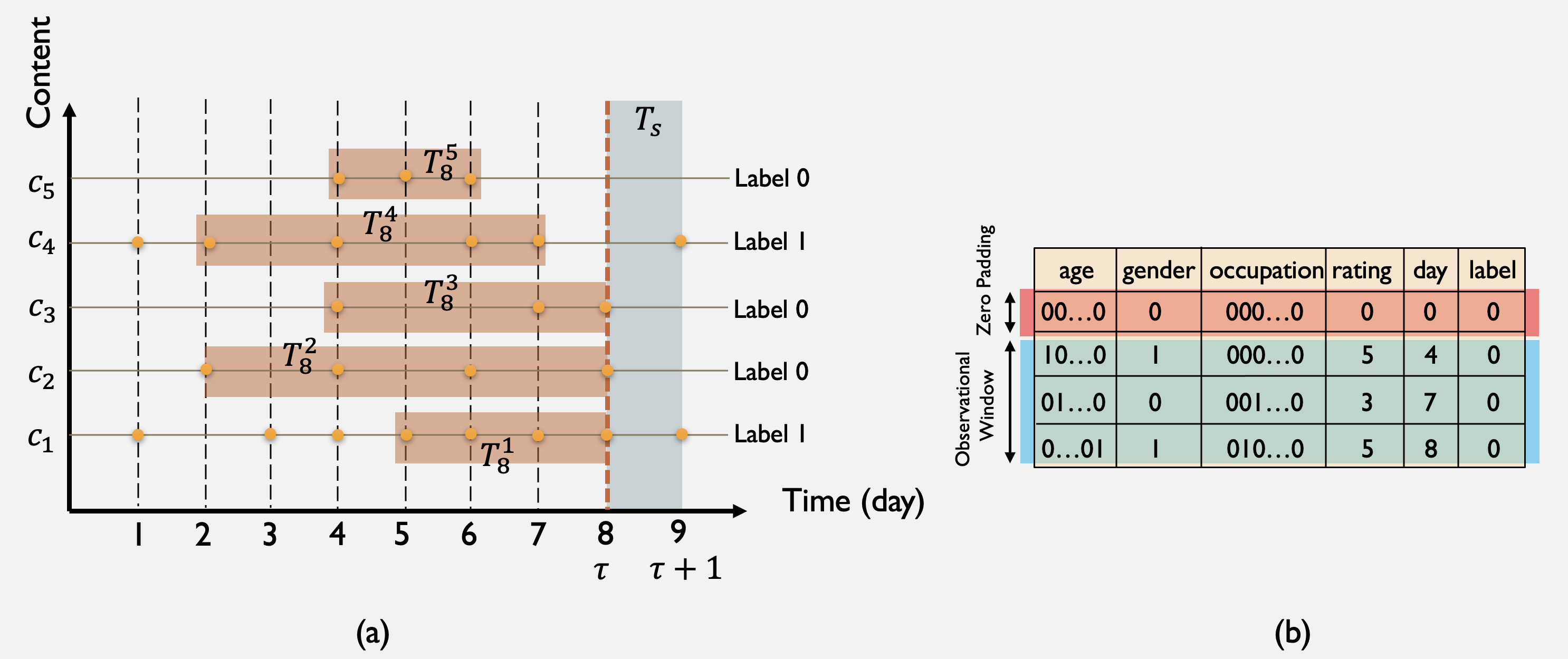}
\caption{\footnotesize (a) An illustrative example of request patterns of contents to construct input samples and their labels. (b) Zero padding technique is used for contents with the number of requests less than $N_o$.} \label{example}
\end{figure*}
To clarify the benefits of the proposed strategy, an illustrative example of constructing input samples and their labels is depicted in Fig.~\ref{example}. In this example, it is assumed that there are $5$ contents $\{c_m\}^{5}_{m=1}$ through the network, and the goal is to update the storage of edge devices at time $\tau + T_s = 9$ using the historical requests of contents up to time $\tau = 8$. Moreover, $N_o = 4$ number of requests for each content are investigated during the observational window. As shown in Fig.~\ref{example}(a), the following cases can occur: 
\begin{itemize}
\item [1.] In the first case, content $c_1$ is requested at time $\tau = 8$. Ending at time $\tau = 8$, the observational window $T_8^1$ is determined in such a way that the number of requests during this observational window is  $N_o= 4$, where $T_8^1 = 3$. Therefore, the requests within $T_8^1 = 3$ ending at $\tau = 8$ are considered to evaluate the popularity of content $c_1$ at time $\tau + T_s = 9$. Consequently, $\{x_k^1\}^{4}_{k=1}$ includes the contextual information of users requesting content $c_1$ and their ratings. As shown in Fig.~\ref{example}(a), since content $c_1$ is requested at time $9$, its label is $\{y_k^1\}^{4}_{k=1} = 1$.
\item [2.] In the second case (i.e., content $c_2$), there is a request at time  $\tau = 8$ and the observational window is $T_8^2 = 6$ for having $N_o= 4$ number of requests within this range. However, since this content is not requested at time $9$, its label is $\{y_k^2\}^{4}_{k=1} = 0$.
\item [3.] In the third case (i.e., content $c_3$), the content is requested at $\tau = 8$, but the number of existing requests from the beginning to $\tau = 8$ is less than $N_o= 4$. Therefore, we use zero padding to create the same length of input samples, shown in Fig~\ref{example}(b), where zero pad will be added at the beginning of the observational window.
\item [4.] In the fourth case, content $c_4$ is not requested at time $\tau = 8$, but there is at least $N_o= 4$ number of requests before $\tau = 8$. Therefore, the last $N_o = 4$ of requests within the observational window $T_o^4$ are used as the input sample.
\item [5.] In the fifth case, content $c_5$ is not requested at time $\tau = 8$, and the number of requests from the beginning to time $\tau = 8$ is less than $N_o= 4$. Similar to the case $3$, the zero padding technique is used to create the input sample. 
\item [6.] Finally, if there is a content with zero requests up to time $\tau = 8$, this content will be removed from the study.
\end{itemize}
This completes  the problem description and system modeling, next, we present the proposed CLSA framework.

\section{Proposed CLSA Framework} \label{CLSA}
\begin{figure*} [t!]
\centering \includegraphics [scale = 0.3] {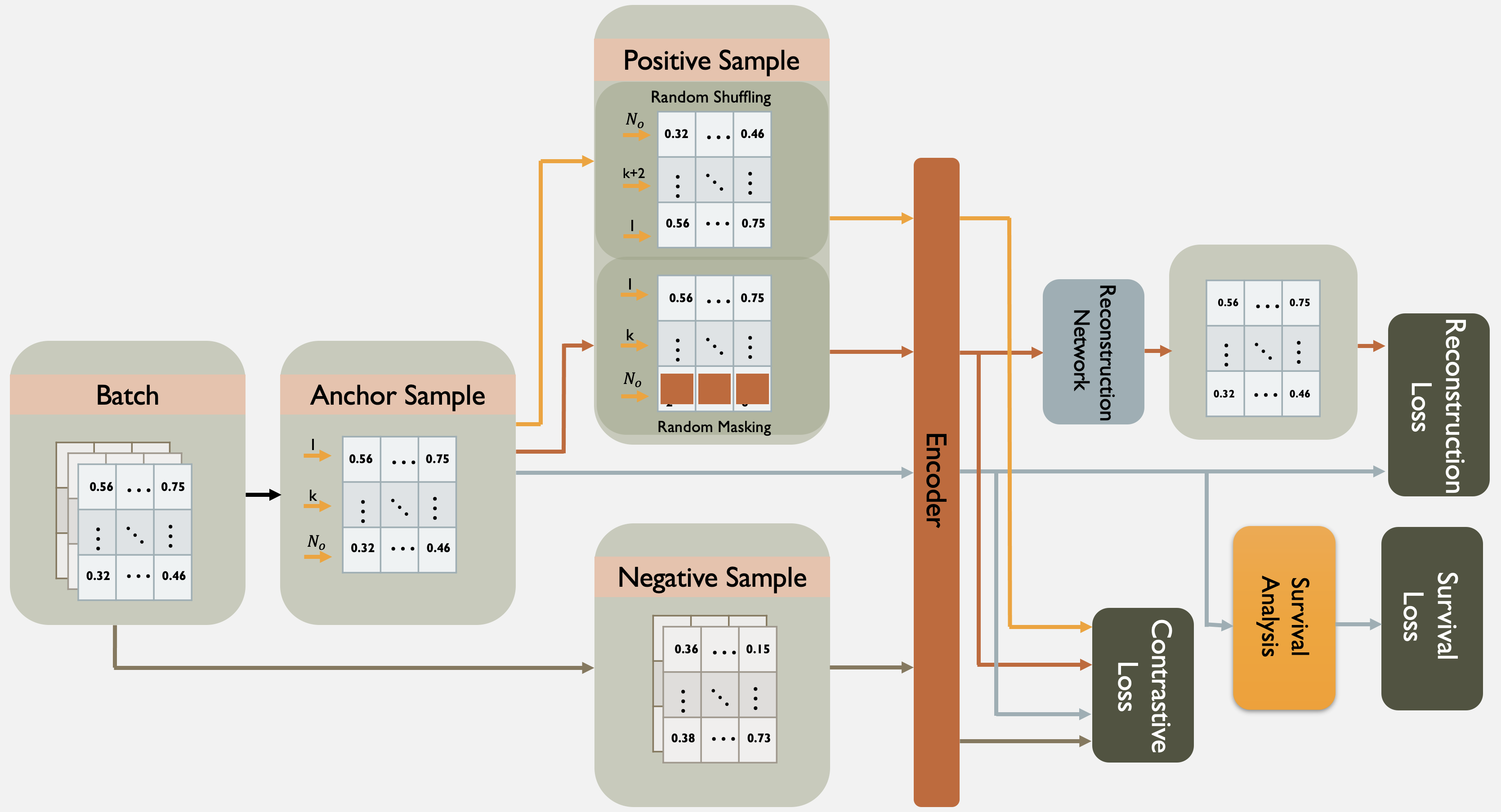}
\caption{\footnotesize The overall perspective of the CLSA architecture.} \label{block}
\end{figure*}

In this section, we present the constituent components of the proposed CLSA framework. As can be seen from Fig.~\ref{block}, the CLSA architecture consists of three modules, i.e., CL, RN, and SA models. The CA network is utilized to capture the spatial and temporal correlations of input samples by converting the longitudinal input data $x_k^m \in \{ d^m_{\tau} \}^{M}_{m=1}$ to the latent representations. The RN block is then used to decode the latent representation. Given a meaningful latent representation, the SA model is used to predict the probability that content $c_m$ will become popular. In the following, first, we briefly introduce the CL block, then, we explain RN and the SA~models.

\subsection{Contrastive Learning (CL) Block} \label{CL}
CL model, as one of the widely used self-supervised learning paradigms in computer vision, has recently been applied to tabular and longitudinal datasets. The fundamental goal of the CL network is to learn a latent representation in which similar samples stay nearby and pairs that are dissimilar to one another move farther apart. As shown in Fig.~\ref{block}, among all samples in each batch, one input sample is considered the anchor sample. Relying on the data augmentation, a positive pair is created using the anchor sample, and other samples in the batch are considered negative samples. In this work, we use the following two data augmentation techniques to generate positive samples (in what follows, for simplicity, we drop subscript $\tau$ wherever there is no ambiguity):
\begin{itemize}
\item \textbf{Masking:} Given the longitudinal data $d_m$ of content $c_m$, several users' information $\{x_k^m\}^{N_o}_{k=1}$ is randomly masked to generate a positive sample for content $c_m$, denoted by $d_m^{(MA)}$. Other historical request patterns of contents in the batch are represented by negative samples.
\item \textbf{Shuffling:} In this type of data augmentation, the positive sample is generated by randomly shuffling the time order of users' information requesting content $c_m$. The shuffled sample is denoted by $d_m^{(SH)}$.
\end{itemize}
Given the augmented and negative samples, a shared encoder is utilized to convert the longitudinal input data to a meaningful latent representation. This representation places contents with similar request patterns closer together and dissimilar contents farther apart. To preserve the temporal correlation of request patterns, an LSTM architecture is used as the encoder, where the latent representation of $x_k^m$ at time $t_{k}^m$, denoted by $h_k^m$, is given by
\begin{equation}
h_k^m = LSTM(x_k^m, h_{k-1}^m),~~~~~~k=1, \ldots, N_o.
\end{equation}
In addition to the anchor sample $d_m$, the shuffled and masked version of $d_m$ are given as the input of the encoder to efficiently learn the latent representation of $d_m$. The shuffled and masked encoded samples are represented by $h_k^{m,(SH)}$, and $h_k^{m,(MA)}$, respectively.

Finally, we utilize the masked and shuffled CL loss functions, denoted by $L_{cl}^{(MA)}$, and $L_{cl}^{(SH)}$, respectively, where the shuffled/masked learned representation is considered the positive sample, and other representations in the batch are considered the negative ones, as follows
\begin{equation}
L_{cl}^{(MA)} = - \sum \limits_{m=1}^{M} \log \dfrac{\exp \big(h_{N_o}^m (h_{N_o}^{m,(MA)})^T\big)}{\sum \limits_{j=1, j \neq m}^{M} \exp \big(h_{N_o}^m (h_{N_o}^{j,(MA)})^T \big)},
\end{equation}
\begin{equation}
L_{cl}^{(SH)} = - \sum \limits_{m=1}^{M} \log \dfrac{\exp\big(h_{N_o}^m (h_{N_o}^{m,(SH)})^T\big)}{\sum \limits_{j=1, j \neq m}^{M} \exp\big(h_{N_o}^m (h_{N_o}^{j,(SH)})^T \big) }, 
\end{equation}
where $(.)^T$ is the transpose function, and the total CL loss is $L_{cl} = L_{cl}^{(MA)} + L_{cl}^{(SH)}$. This completes the presentation of the CL component of the CLSA architecture. Next, we present the reconstruction network.
\subsection{Reconstruction Network (RN) Block} \label{}
To learn a better latent representation, the masked version of $x_k^m$ is regenerated from the encoded masked input sample. Following Reference~\cite{Zhu2017} and to preserve the temporal information of $\{x_k^m\}^{N_o}_{k=1}$, the Time-LSTM2 is utilized as the decoder, where the regenerated sample is given by
\begin{equation}
\overline{x}_k^m = Time-LSTM2(H_k^m),
\end{equation}
where $\overline{x}_k^m$ is the decoded version of $x_k^m$, and $H_k^m = [(h_1^{m,(MA)}, t_2^m - t_1^m), (h_2^{m,(MA)}, t_3^m - t_2^m), \ldots, (h_k^{m,(MA)}, t_{k+1}^m - t_k^m)]$ with $k \in \{1, \ldots, N_o\}$. Term $h_k^{m,(MA)}$ is the masked encoded sample at time $t_k^m$, and $t_k^m- t_{k-1}^m$ is the time difference between two consecutive requests of content $m$. By minimizing the difference between the original input sample $x_k^m$ and the decoded one $\overline{x}_k^m$, our goal is to provide a better latent representation. Therefore, the RN loss, denoted by $L_{re}$, is given by
\begin{equation}
L_{rn} = \sum_{k=1}^{N_o} ||\overline{x}_k^m - x_k^m ||^2.
\end{equation}

\subsection{Survival Analysis (SA) Block} \label{}
We utilize a SA model to capture a meaningful relationship between the longitudinal and contextual information of users requesting content $c_m$ and the probability of getting popular in the future. To this end, we use an MLP network, where the input of this model is the latent representation $\{h_{N_o}^{m}\}^{M}_{m=1}$, and the output is the probability that content $c_m$ gets popular at time $t$, denoted by $p_{t}^m = P(t |h_{N_o}^{m})$, during the total time window, denoted by $T_{total}$, where $T_{total} = T_s + \max \limits_{1\leq m \leq M}T_o^m$. We use the softmax function as the activation function of the output layer of the MLP network to compute $p_{t}^m$. Consequently, the estimated Cumulative Incidence Function (CIF) for content $c_m$ at time $\tau + T_s$ is calculated as follows
\begin{equation}
\overline{F}^m (\tau + T_s | h_{N_o}^{m}) =  \dfrac{\sum \limits_{\tau \leq t \leq \tau + T_s} p_{t}^m}{1 - \sum \limits_{t \leq t_{N_o}^{m}} p_{t}^m},
\end{equation}
where $\overline{F}^m (\tau + T_s | h_{N_o}^{m})$ indicates the probability that content $m$ has gained popularity up to time $ \tau + T_s  $. Finally, the negative log-likelihood is used as the loss function for the SA network, given by
\begin{eqnarray}
L_{sa}= - \sum \limits_{m=1}^{M}  \log \big( 1-  \overline{F}^m (\tau^m | h_{T_m}^{m})  \big) .
\end{eqnarray}
Finally, $\{p_{t}^m\}^{M}_{m=1}$ are sorted in the descending order to identify the Top-$K$ popular contents. 

\section{Simulation Results} \label{sec:4}

\subsection{Experimental Configurations} 
To investigate the effectiveness of the proposed CLSA architecture, a UAV-aided MEC network, consisting of $4$ terrestrial and $2$ aerial caching nodes with $943$ users and $1682$ multimedia contents. Following the common assumption~\cite{Hajiakhondi2019}, the size of the storage capacity of caching nodes is $10\%$ of the total contents, where the size of all multimedia contents is the same. Using the five-fold cross-validation strategy, $80\%$ and $20\%$ samples are used as the training dataset and the test dataset, respectively. Adam optimizer was employed to train the model, where betas are ($0.9, 0.999$) and the weight decay is set to $1e-7$. Moreover, the $l_2$ regularization was set to $1e-3$ to avoid over-fitting. We trained the proposed CLSA architecture by minimizing the total loss function, denoted by $L_{total}$, obtained as follows
\begin{equation}
L_{total} = \omega_{cl} L_{cl} + \omega_{rn} L_{rn} + \omega_{sa} L_{sa},
\end{equation}
where $\omega_{cl}$, $\omega_{rn}$, and $\omega_{sa}$ represent the weight of CL, RN, and SA blocks, respectively, where the summation of them is one. In the following, the details of each learning block are described:
\begin{itemize}
\item \textbf{Encoder:} An LSTM network is used as the encoder, with Rectified Linear Unit (ReLU) activation and sigmoid as the recurrent activation function, where the output size of this block is denoted by $D_{\text{E}}$. Then, the output is given to an MLP with three layers, with the size of $\alpha_l D_{\text{FI}}$, where $l \in \{1,2,3\}$. Term $D_{\text{FI}}$ represents the feature dimension of the input sample, which is $28$ in this work, and the hyperparameter $\alpha_l$, with $l \in \{1,2,3\}$ is set to $\alpha_1 = 1, \alpha_2 = 3$, and $\alpha_3 = 5$.

\item \textbf{Decoder:} There is a decoder in the RN block, performing based on the Time-LSTM2, where the recurrent activation function is sigmoid and the general activation is Tanh. The output size of the Time-LSTM2 network is denoted by $D_{\text{D}}$.

\item \textbf{MLP Network:} Using batch-normalization technique, this network consists of three dense layers with the same size of $D_{\text{M}}$ and the Exponential Linear Unit (ELU) activation function. There are more $5$ dense layers after that, where the size of each layer is $(5,3,2,1,1) \times T_{total}$. The activation function of the last layer is softmax, while the rest is ReLU.
\end{itemize}

\subsection{Effectiveness of the CLSA Architecture}
To evaluate the performance of the proposed CLSA architecture, we first consider different variants of the CLSA model by changing different hyperparameters, such as the batch size, $D_{\text{E}}$,  $D_{\text{D}}$, $D_{\text{M}}$, and the learning rate, denoted by $lr$. According to the information provided in Table~\ref{table1}, five different models are defined. Moreover, we investigate the effect of the number of requests studied in the observational window, denoted by $N_o$, on the classification accuracy. As it can be seen from Table~\ref{table2}, decreasing the batch size from $512$ to $256$ (Model $1$ and $2$) improves the classification accuracy over all folds with different $N_o$. Comparing Models $2$ and $3$, it can be seen that reducing the encoder and decoder dimensions from $512$ to $256$ decreases the classification accuracy. Similarly, reducing $D_{\text{M}}$ from $128$ to $32$ results in degrading the accuracy (Model $2$ and Model $4$). Finally, by comparing Model $1$ and Model $5$, it is evident that $lr = 1e-3$ outperforms  $lr = 1e-4$.

Moreover, we investigate the effect of $N_o$ on the classification accuracy over $5$ models. According to the results provided in Table~\ref{table2}, increasing the number of requests studied over an observational window provides more information about the behavior of users' interests in the past, improving the classification accuracy. Accordingly, from the aspect of classification accuracy, it can be seen that Model $2$ with $N_o=20$ outperforms other variants. For this reason, we have selected this model to conduct further research. Table~\ref{table3} demonstrates the precision, recall, and F1-score for Model $2$ with $N_o=20$ over $5$ folds and their average values. According to the information provided in Table~\ref{table3}, the high value of the aforementioned parameters illustrates the effectiveness of the CLSA architecture. 

\begin{figure}[t!]
\centering
\vspace{-.05in}
\includegraphics[scale=0.6]{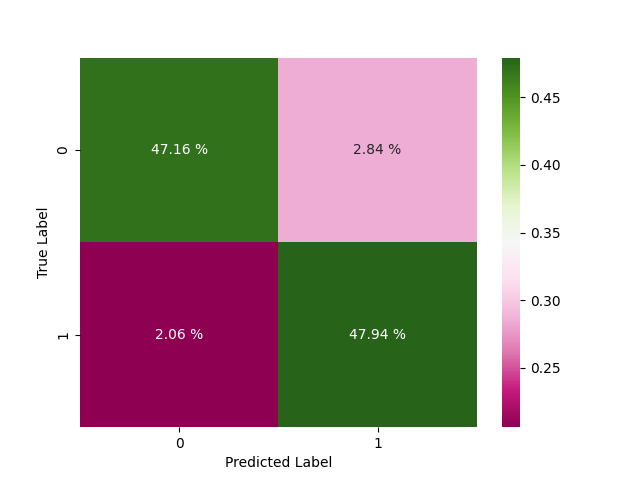}
\caption{\footnotesize Confusion matrix of the proposed CLSA architecture (Model 2, $N_o=20$). }\label{confusion}
\end{figure}

Fig.~\ref{confusion} represents the confusion matrix of the Model $2$ with $N_o=20$. It should be noted that a challenging issue encountered in the Movielens dataset pertains to imbalanced data, which arises due to a vast proportion of multimedia contents being unpopular. To address this problem, we implemented the random oversampling technique to increase the number of popular samples. As depicted in Fig.~\ref{confusion}, there is a misclassification rate of $2.84\%$ for unpopular contents (labeled as 0) being incorrectly classified as popular contents, which leads to the wastage of storage capacity on edge devices. Similarly, a misclassification rate of $2.06\%$ for popular contents being classified as unpopular contents can result in failure to place highly requested contents on edge devices.

\begin{figure}[t!]
\centering
\vspace{-.05in}
\includegraphics[scale=0.4]{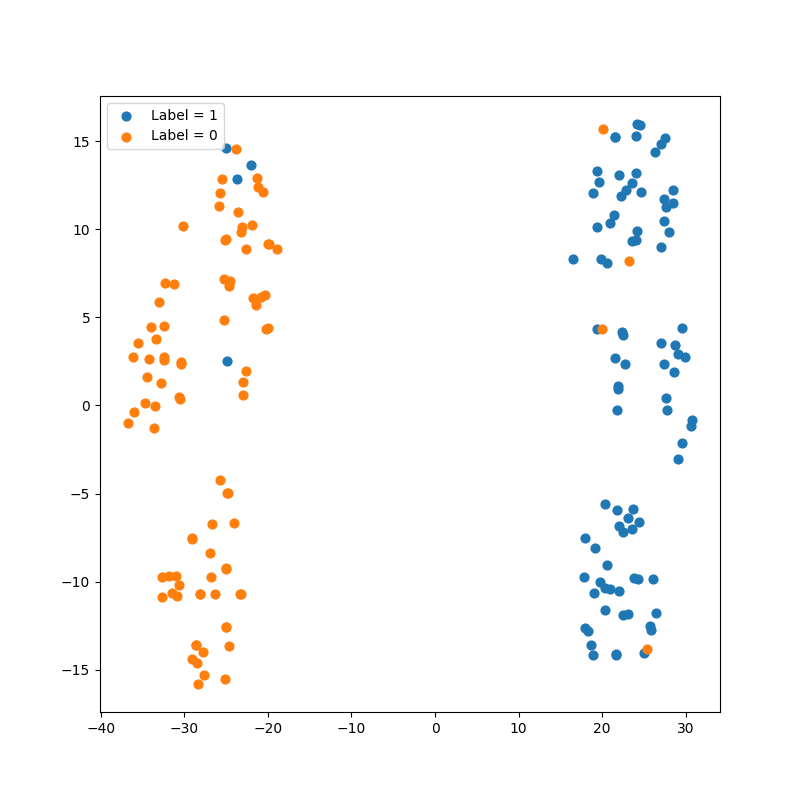}
\caption{\footnotesize The embedded space of the latent representation of popular/unpopular contents using the TSNE technique.}\label{pca}
\end{figure}

Moreover, we utilize the T-distributed Stochastic Neighbor Embedding (TSNE) method~\cite{Cieslak2020} to evaluate the efficiency of the CL block in generating latent representations for discriminating between popular and unpopular contents. For instance, in Fig.~\ref{pca}, the latent representation of a test set from one of the $5$ fold cross-validation experiments is employed to illustrate the embedded space of popular and unpopular contents. As shown in Fig.\ref{pca}, the embedded space of popular and unpopular contents are distinguishable, indicating that the CL network has been effectively trained.

\begin{table}[t]
\centering
\renewcommand\arraystretch{2}
\caption{\small Variants of the CLSA architecture.}
\label{table1}
{\begin{tabular}{   c | c c c  c  c  c | c }
\hline
\hline
\textbf{Model ID}
& \textbf{Batch Size}
& \textbf{$D_{\text{E}}$}
& \textbf{$D_{\text{D}}$}
& \textbf{$D_{\text{M}}$}
& \textbf{$lr$}
\\
\hline
\textbf{1}
& 512
& 512
& 512
& 128
& $1e-3$
\\
\textbf{2}
& 256
& 512
& 512
& 128
& $1e-3$
\\
\textbf{3}
& 256
& 256
& 256
& 128
& $1e-3$
\\
\textbf{4}
& 256
& 512
& 512
& 32
& $1e-3$
\\
\textbf{5}
& 256
& 512
& 512
& 128
& $1e-4$
\\
\hline
\end{tabular}}
\end{table}

\begin{table*}[t]
\centering
\renewcommand\arraystretch{2}
\caption{\small 5 fold cross-validation accuracy $\pm$ standard deviation for different variants of the proposed CLSA architecture using different window sizes (10, 15, and 20 days).}
\label{table2}
{\begin{tabular}{ |  c | c | c |  c | c | c | c | c |}
\hline
\textbf{$N_o$}
&\textbf{Model ID} 
&\textbf{Fold 1}
& \textbf{Fold 2}
& \textbf{Fold 3}
& \textbf{Fold 4}
& \textbf{Fold 5}
& \textbf{Average}
\\
\hline 
\multirow{ 5}{*}{10} &
\textbf{1} & $0.850 \pm	0.026 $ & $0.836 \pm 0.037$& $	0.837 \pm 0.024$ &$	0.875	\pm 0.022 $& $0.831 \pm 0.050$ & $	0.846 \pm 0.035$
\\
&
\textbf{2} &
$0.892 \pm 0.031$ & $0.909 \pm 0.011$ & $	0.885 \pm 0.031$ & $	0.905 \pm 0.016$ &$	0.892 \pm 0.021$ & $	0.896 \pm 0.024$
\\
&\textbf{3}
& $0.853 \pm 0.023$ & $	0.829 \pm 0.030$ & $	0.826 \pm 0.025$ &$	0.848 \pm 0.036$ &$	0.815 \pm 0.029$ & $0.834 \pm 0.031$
\\
&\textbf{4}
&$0.864 \pm 0.018$ &$	0.826 \pm 0.012$ & $	0.835 \pm 0.015$ &$	0.849 \pm 0.018$ & $0.844 \pm 0.031$ &$	0.843 \pm 0.023$
\\
&\textbf{5}
& $0.866 \pm 0.012$ &$	0.831 \pm 0.015$ & $	0.834 \pm 0.010$ & $	0.857 \pm 0.012$ & $	0.859 \pm 0.030$ & $	0.849 \pm 0.022$
\\
\hline
\multirow{ 5}{*}{15} &
\textbf{1}&
$0.890 \pm 0.031$ & $0.868 \pm	0.032$ & $0.843 \pm 0.052$ & $	0.861 \pm	0.027$ & $0.861 \pm 0.050$ &$0.865 \pm 0.041$
\\
&
\textbf{2}
& $0.949 \pm 0.019$ & $	0.940 \pm 0.007$ & $	0.934 \pm 0.010$ & $	0.937 \pm 0.005$ & $	0.932 \pm 0.012$ & $	0.938 \pm 0.012$
\\
&\textbf{3}
&  $0.861 \pm 0.022$ & $	0.836 \pm 0.024$ &$	0.842 \pm 0.031$ & $	0.861 \pm 0.018$ &$	0.843 \pm 0.033$ & $	0.849 \pm 0.027$
\\
&\textbf{4}
& $0.867 \pm 0.026$ &$	0.832 \pm 0.025$ & $0.835 \pm 0.024$ & $0.868 \pm 0.023$ 	&$0.850 \pm 0.029$ 	&$ 0.850 \pm 0.029$
\\
&\textbf{5}
& $0.864 \pm 0.009$ & $	0.826 \pm 0.010$ & $	0.823 \pm 0.012$ & $	0.850 \pm 0.021$ & $0.870 \pm 0.019$ & $	0.847 \pm 0.024$
\\
\hline
\multirow{ 5}{*}{20} &
\textbf{1} & $0.878\pm0.037$ & $0.881 \pm0.041$ & $0.859 \pm 0.050$&	$0.868 \pm 0.058	$ &	$0.855 \pm 0.023$ & $0.868 \pm 0.043$
\\
&\textbf{2}
&$0.959 \pm 0.003$& $0.942 \pm 0.006$ &$0.941 \pm 0.006$ & $0.946  \pm 0.002 $ & $	0.965 \pm 0.003 $ & $ \textbf{0.951} \pm \textbf{0.010}$
\\
&\textbf{3}
& $0.866 \pm 0.033$ &$	0.875 \pm 0.032$ & $	0.876 \pm 0.019$ & $	0.854 \pm 0.024$ & $ 	0.845 \pm 0.032$ & $	0.863 \pm 0.029 $ 
\\
&\textbf{4}
& $0.858 \pm 0.016$ & $	0.845 \pm 0.039 $ &$0.834 \pm 0.013 $ & $	0.869 \pm 0.034$ & $	0.829 \pm 0.065$ & $	0.847 \pm 0.039$
\\
&\textbf{5}
&$0.871 \pm	0.009 $ &$0.828 \pm 0.013$ & $	0.830 \pm 0.022$ & $	0.859 \pm 0.009$ &$	0.878 \pm 0.006$ & $	0.853 \pm 0.024$
\\
\hline
\end{tabular}}
\end{table*}

\begin{table*}[t]
\centering
\renewcommand\arraystretch{2}
\caption{\small Precision, recall, and F1-score for two classes (i.e., popular (class 1) and unpopular (class 0)) for Model 2 with $N_o = 20$ using $5$ fold cross-validation.}
\label{table3}
{\begin{tabular}{ | c | c | c | c | c | c|  c | c |}
\hline
&\textbf{Class} 
&\textbf{Fold 1}
& \textbf{Fold 2}
& \textbf{Fold 3}
& \textbf{Fold 4}
& \textbf{Fold 5}
& \textbf{Average}
\\
\hline
\multirow{ 2}{*}{Precision} &
\textbf{0} & $0.973 \pm 0.012$ & $	0.934 \pm 0.019$ & $0.911 \pm 0.005$ & $	0.928 \pm 0.005$ & $	0.956 \pm 0.010$ & $	0.941 \pm 0.024$
\\
& \textbf{1}
&  $0.946 \pm 0.009$ & $	0.938 \pm 0.019$ & $	0.976 \pm 0.011$ & $	0.951 \pm 0.006$ & $	0.973 \pm 0.005$ & $	0.957 \pm 0.018$
\\
\hline
\multirow{ 2}{*}{Recall} &
\textbf{0} & $0.945 \pm 0.010$ & $	0.937 \pm 0.022$ & $	0.978 \pm 0.010$ & $	0.952 \pm 0.006$ & $	0.974 \pm 0.006$ & $	0.957 \pm 0.019$
\\ 
& \textbf{1}
&  $0.974 \pm 0.012$ & $	0.934 \pm 0.021$ & $	0.904 \pm 0.006$ & $	0.926 \pm 0.006$ & $	0.956 \pm 0.010$ & $	0.939 \pm 0.027$
\\
\hline
\multirow{ 2}{*}{F1-score} &
\textbf{0} & $0.959 \pm 0.004$ & $	0.935 \pm 0.004$ & $	0.943 \pm 0.006$ & $	0.940 \pm 0.005$ & $	0.965 \pm 0.002$ & $	0.948 \pm 0.012$
\\
&
\textbf{1}
&  $0.960 \pm 0.003$ & $0.935 \pm 0.003$ & $	0.939 \pm 0.003$ &$	0.938 \pm 0.005$ & $	0.964 \pm 0.003$ & $	0.947 \pm 0.013$
\\
\hline
\end{tabular}}
\end{table*}
\setlength{\textfloatsep}{0pt}

\vspace{-.05in}
\subsection{Ablation Study}
To further investigate the effectiveness of the proposed CLSA architecture, we conduct an ablation study, where different variants of the CLSA model are introduced in Table~\ref{table4}. We evaluate the significance of each block in the CLSA architecture on classification accuracy by analyzing Models $L1$ to $L7$. For instance, the results presented in Table~\ref{table4} demonstrate that the CL block alone (Model $L1$) achieves an accuracy of approximately $85\%$, underscoring the importance of the other two blocks in attaining higher performance (Model $L7$ with about $95 \%$ accuracy). Consequently, it is evident that the presence of all three blocks in the CLSA architecture is of paramount importance, where the optimal loss weights of each block are represented in Model $L7$.

\subsection{Effectiveness of the CLSA Architecture}
In this subsection, we compare the performance of the proposed CLSA architecture with the following state-of-the-art:
%

\begin{table}[t]
\centering
\renewcommand\arraystretch{2}
\caption{\small The accuracy of the proposed CLSA architecture using different loss weights.}
\label{table4}
{\begin{tabular}{ |  c | c   c  c | c |}
\hline
\textbf{Model ID} 
&\textbf{$\omega_{cl}$}
& \textbf{$\omega_{rn}$}
& \textbf{$\omega_{sa}$}
& \textbf{Average Accuracy $\pm$ STD}
\\
\hline
$L1$
& 1.0
& 0.0
& 0.0
&  $0.851 \pm 0.026$
\\
$L2$
& 0.0
& 1.0
& 0.0
& $0.850 \pm 0.022 $
\\
$L3$
& 0.0
& 0.0
& 1.0
& $0.903 \pm 0.029 $
\\
$L4$
& 0.5
& 0.5
& 0.0
& $ 0.859 \pm 0.034 $
\\
$L5$
& 0.0
& 0.5
& 0.5
& $ 0.888 \pm 0.025$
\\
$L6$
& 0.5
& 0.0
& 0.5
& $ 0.936 \pm 0.013$
\\
$L7$
& 0.3
& 0.2
& 0.5
& $\textbf{0.951} \pm \textbf{0.010}$
\\
\hline
\end{tabular}}
\end{table}

\begin{itemize}
\item \textbf{Transformer-based Edge Caching (TEDGE) Scheme}~\cite{Hajiakhondi2021_ICC}, which is based on a simple ViT architecture acting as a multi-label classification model with the aim of predicting the Top-K popular contents in the upcoming time. To capture the spatial correlation of contents, $2$D images of historical requests pattern of contents were created, where the number of columns and rows of this image were corresponding to the number of contents and the number of historical requests for each content, respectively. The size of the input sample, therefore, significantly increases to capture as much spatial correlation as possible.

\item \textbf{Multiple-model Transformer-based Edge Caching (MTEC)}~\cite{Hajiakhondi2022_IoT} consists of two parallel multi-channel Transformer networks with a dense layer as the fusion layer. Similarly, the output of the model is Top-K popular contents, while it first predicted the request patterns of contents in the future. The input sample is $1$D historical requests patterns of contents. To capture the spatial correlation of contents, the multi-channel Transformer networks were employed, where the sequential request pattern of each content is given to a channel of the Transformer model. 

\item \textbf{Vision Transformers with Cross Attention (ViT-CAT)}~\cite{Hajiakhondi2022_Icassp}, consisting of two parallel ViT networks with different patching techniques, with a cross attention mechanism as the fusion layer. The input and output of the network are similar to the TEDGE caching scheme. 

\item \textbf{Self-Supervised Contrastive Learning using Random Feature Corruption (SCARF)}~\cite{Bahri2022}, which was applied on tabular datasets. We modified it for the popularity prediction task, where the input sample is the $1$D historical requests of content, without any need for labeling. SCARF attempts to learn the latent representation of contents to classify them as popular and unpopular contents using random feature corruption.

\item \textbf{Deep Learning-based Content Caching (DLCC)}~\cite{Bhandari2021} used CNN to predict the popularity of contents and used the RL model for the content placement phase.
\end{itemize}
We compare the proposed CLSA framework with the aforementioned baselines from the classification accuracy perspective in Table~\ref{table5}. As shown in Table~\ref{table5}, the proposed CLSA architecture outperforms other baselines, while there is no need to create large input samples to capture the spatial correlation of contents.

\begin{table}[t]
\centering
\renewcommand\arraystretch{2}
\caption{\small Comparison with state-of-the-art based on the classification accuracy.}
\label{table5}
{\begin{tabular}{ |  c | c |}
\hline
\textbf{Model} 
& \textbf{Accuracy}
\\
\hline
TEDGE (ViT)~\cite{Hajiakhondi2021_ICC}
&  $93.72 \%$
\\
MTEC~\cite{Hajiakhondi2022_IoT}
& $94.13 \%$
\\
ViT-CAT~\cite{Hajiakhondi2022_Icassp}
& $94.84 \%$
\\
SCARF~\cite{Bahri2022}
& $87.17 \%$
\\
DLCC~\cite{Bhandari2021}
& $92.81 \%$
\\
Proposed CLSA
&$\textbf{95.10 \%}$
\\
\hline
\end{tabular}}
\end{table}

\begin{figure}[t!]
\centering
\includegraphics[scale=0.38]{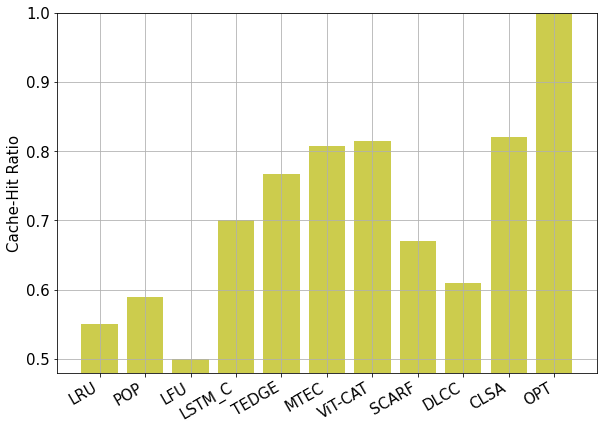}
\caption{\footnotesize Comparison with state-of-the-art based on the cache-hit ratio.}\label{cache}
\end{figure}
Finally, we compare the performance of the proposed CLSA architecture with other baselines in terms of the cache-hit ratio. Note that the cache-hit ratio is commonly used in MEC networks to evaluate the effectiveness of the popularity prediction framework. This metric shows the number of requests that are being handled by caching nodes versus the overall number of requests made throughout the network. As shown in Fig.~\ref{cache}, there are other baselines in addition to the aforementioned schemes, including Least Recently Used (LRU)~\cite{Giovanidis2016}, Least Frequently Used (LFU)~\cite{Giovanidis2016}, PopCaching~\cite{Li2016}, and LSTM-C~\cite{Zhang2019}. As depicted in Fig.~\ref{cache}, the optimal approach~\cite{Zhang2019} is a caching scheme where caching nodes handle all requests throughout the network, which is not feasible in real-world scenarios. Based on the results presented in Fig.~\ref{cache}, the proposed CLSA architecture achieves the highest cache-hit ratio when compared to other baselines.

\vspace{-.1in}
\section{Conclusion}\label{sec:5}
In this paper, we developed the Contrastive Learning-based Survival Analysis (CLSA) popularity prediction framework with the application to the Mobile Edge Caching (MEC) networks. To learn the temporal information of sequential requests, the proposed architecture utilized a self-supervised Contrastive Learning (CL) model that employed a Long Short-Term Memory (LSTM) network as the encoder. Unlike existing research works that used multiple contents' historical request patterns simultaneously to capture spatial dependency, the proposed CLSA architecture used the CL network, eliminating the need for large input samples. The output of the architecture was the probabilities of each content's future popularity, which were sorted in descending order to determine the Top-K popular contents. Employing the CL network not only reduced the input size significantly but also enhanced the scalability of the learning model. Adding new content to the network does not require recreating input samples, making it easier to scale up the popularity prediction framework. Simulation results illustrated that the proposed CLSA architecture improved the cache-hit ratio and classification accuracy when compared to its state-of-the-art. Going forward, several directions deserve further investigation. One area of interest is the implementation of a Contrastive Learning-based Graph Neural Network (GNN) to establish a meaningful connection between different users and their interests. Another important aspect is the role of the augmentation scheme in CL performance. Therefore, developing an effective augmentation method to create more relevant positive samples is essential for improving the effectiveness of the popularity prediction framework.

\end{document}